\documentclass[letterpaper]{article} 
\usepackage{aaai25}  
\usepackage{times}  
\usepackage{helvet}  
\usepackage{courier}  
\usepackage[hyphens]{url}  
\usepackage{graphicx} 
\usepackage{natbib}  
\usepackage{caption} 
\usepackage{algorithmic}
\usepackage{threeparttable}
\usepackage{diagbox}
\usepackage{multirow}
\usepackage{booktabs} 
\usepackage{subcaption} 
\usepackage{latexsym, amsmath, amssymb, amsfonts, mathrsfs}
\usepackage{xspace}
\usepackage[T1]{fontenc}
\usepackage{float}
\usepackage{color}
\usepackage{url}
\usepackage[ruled,linesnumbered,noresetcount,vlined]{algorithm2e}
\usepackage{todonotes}
\usepackage{picinpar}
\usepackage{multirow}
\usepackage{subcaption}
\usepackage{amsthm}
\usepackage{booktabs} 
\usepackage{diagbox}
\allowdisplaybreaks
\usepackage{pgfplots}

\newcommand{\fml}[1]{\mathcal{#1}}

\newcommand{\beitemize}{\begin{list}{$\bullet$}{\topsep=1.5pt \parsep=0pt \itemsep=1pt \leftmargin=1em }} 
\newcommand{\enitemize}{\end{list}}

\newcommand{\beenumerate}{\hspace{-0.5in} \begin{enumerate}\topsep=1pt \parsep=0pt \itemsep=-3pt} \newcommand{\enenumerate}{\end{enumerate}}

\newcommand{\belist}{\begin{list}{$\bullet$}{\topsep=1.5pt \parsep=0.5pt \itemsep=1pt \leftmargin=2.25em \labelwidth=1.0em \labelsep=0.5em \partopsep=1.5pt}} 
\newcommand{\enlist}{\end{list}}

\newcommand{\argmax}{\operatornamewithlimits{argmax}}

\newtheorem{definition}{{\bf Definition}}

\newtheorem{example}{{\bf Example}}

\newboolean{includeMemo}
\setboolean{includeMemo}{true} 

\newcommand{\memoside}[1]{\ifthenelse{\boolean{includeMemo}}{\todo[caption={},color=green!20!]{{\footnotesize #1}}}}
\newcommand{\memo}[1]{\ifthenelse{\boolean{includeMemo}}{\todo[inline,caption={},color=green!20!]{#1}}}
\newcommand{\memob}[1]{\ifthenelse{\boolean{includeMemo}}{\todo[inline,caption={},color=blue!20!]{#1}}}

\newcommand{\xhdr}[1]{\vspace{5pt}\noindent\textbf{#1 }}
\newcommand{\ignore}[1]{}

\newcommand{\squishlist}{
\begin{list}{{{\small{$\bullet$}}}}
{\setlength{\itemsep}{3pt}      
\setlength{\parsep}{3pt}
\setlength{\topsep}{3pt}       
\setlength{\partopsep}{3pt}
\setlength{\leftmargin}{1em} 
\setlength{\labelwidth}{1em}
\setlength{\labelsep}{0.5em} } }
\newcommand{\squishend}{  \end{list}}

\newcommand{\squishenum}{
\begin{list}{$\bullet$}{ 
    \setlength{\itemsep}{1pt}
    \setlength{\parsep}{0pt}
    \setlength{\topsep}{1.5pt}
    \setlength{\partopsep}{0pt}
    \setlength{\leftmargin}{2em}
    \setlength{\labelwidth}{1.5em}
    \setlength{\labelsep}{0.5em} } }

\usepackage{newfloat}
\usepackage{listings}
\DeclareCaptionStyle{ruled}{labelfont=normalfont,labelsep=colon,strut=off} 
\floatstyle{ruled}
\newfloat{listing}{tb}{lst}{}
\floatname{listing}{Listing}
\title{Does Your AI Agent Get You? A Personalizable Framework for Approximating Human Models from Argumentation-based Dialogue Traces}
\author{
    Yinxu Tang,
    Stylianos Loukas Vasileiou,
    William Yeoh
}
\affiliations{

    Washington University in St. Louis\\
    \{t.yinxu, v.stylianos, wyeoh\}@wustl.edu
}

\usepackage{bibentry}

\begin{document}

\maketitle

\begin{abstract}
Explainable AI is increasingly employing argumentation methods to facilitate interactive explanations between AI agents and human users.
While existing approaches typically rely on predetermined human user models, there remains a critical gap in dynamically learning and updating these models during interactions. In this paper, we present a framework that enables AI agents to adapt their understanding of human users through argumentation-based dialogues. Our approach, called \textit{Persona}, draws on prospect theory and integrates a probability weighting function with a Bayesian belief update mechanism that refines a probability distribution over possible human models based on exchanged arguments. Through empirical evaluations with human users in an applied argumentation setting, we demonstrate that Persona effectively captures evolving human beliefs, facilitates personalized interactions, and outperforms state-of-the-art methods.
\end{abstract}

%
\begin{links}
\link{Resources}{https://github.com/YODA-Lab/Persona}
\end{links}

\section{Introduction}

As AI systems become more integrated into real-world applications, the need for transparency and trust in human-AI interactions grows. 
Explainable AI (XAI) addresses this need by focusing on generating understandable explanations for human users that foster trust and accountability \citep{gunning2019darpa}. 
A key paradigm within XAI is argumentation \citep{vcyras2021argumentative}, which enables interactive, dialogue-based explanation processes between AI agents and human users. These processes offer improved clarity and foster stronger human-AI interactions.

A core assumption in most existing argumentation-based XAI work is that the AI agent has a static, deterministic model of the human user that it uses in its deliberative processes. While assuming the AI agent has access to an a-priori human model has its advantages \cite{sreedharan2021foundations,vas21}, this approach often falls short of capturing the intricate complexities of real-world interactions. It is not only likely that humans hold beliefs at different levels of granularity and with varying degrees of certainty, but also that their beliefs evolve dynamically over time. Such simplifications can lead to significant misalignments between AI agents and human users \citep{russell2019humancompatible}, as the AI agent might base its decisions or explanations on an inaccurate or incomplete understanding of the human.

As a step towards addressing this issue, in this paper, we propose a novel approach that enables AI agents to adapt their decisions and explanations based on a dynamic understanding of human mental states. Our method represents human models as probability distributions that are continuously refined through ongoing argumentative interactions, building upon established frameworks in computational argumentation \citep{gordon1994,parsons2003properties,prakken2006formal,hunter2015modelling,hunter2016persuasion,Rago_23,vasileiou_DR}. Our proposed framework, \textit{Personalized Human Model Approximations} (Persona), integrates two key components to achieve this goal. First, it employs a \emph{Bayesian belief update} mechanism that systematically refines the human model based on observed interaction patterns. Second, it incorporates a probability weighting function derived from \emph{prospect theory} \citep{tversky1992advances}, which accounts for human tendencies to overweight low probabilities and underweight high probabilities in decision-making contexts. This dual approach enables Persona to offer personalized interactions by capturing individual differences in how users evaluate probabilistic information during argumentative exchanges.

Furthermore, we conduct an extensive evaluation of our approach using real argumentation-based dialogue traces collected via a human-subject study. Our results show that Persona not only effectively captures and updates human models, but it also outperforms existing state-of-the-art argumentation-based methods.

The main contributions of this paper are as follows:
\squishlist
    \item We introduce \textit{Persona}, a novel framework for approximating and updating a probabilistic human model through argumentation-based dialogue traces. Our framework incorporates a prospect-theory-inspired probability weighting function with a Bayesian belief update mechanism.

    \item We conduct a human-subject study on an argumentation-based dialogue scenario and collect dialogue traces involving human users. We empirically evaluate the effectiveness of our approach on these traces and demonstrate its ability to capture evolving human models and facilitate personalized interactions, while also outperforming state-of-the-art methods.
\squishend

\section{Related Work}

\subsection{Argumentation-based Dialogues}

According to the influential work by \citet{walton1995commitment}, dialogues can be categorized based on the knowledge of the participants, the objectives they wish to achieve through the dialogue, and the rules that are intended to govern the dialogue. Contextual to each type, each dialogue revolves around a topic, typically a proposition, that is the subject matter of discussion. Related dialogue types include:~Persuasion \citep{gordon1994,prakken2006formal}, where an agent attempts to convince another agent to accept a proposition they initially do not hold; information-seeking \citep{parsons2003properties,fan2012agent}, where an agent seeks to obtain information from another agent believed to possess it; and inquiry \citep{hitchcock2017some,black2009inquiry}, where two agents collaborate to find a joint proof for a query that neither could prove individually. The advent of argumentative dialogue-based systems \cite{black2021argumentation} illustrates the great potential of argumentation for collaborative decision-making and consensus-building in human-AI interaction settings. However, these approaches often neglect the dynamic nature of belief updating during dialogues.
 
On a similar thread, our work fits well within the literature on argumentation-based explainable AI \cite{fan2015computing,shams2016normative,fan2018on,collins2019towards,budan2020proximity,dennis2022explaining,Rago_23,vasileiou_DR}. While these approaches provide a solid foundation for argumentation-based explanations, they do not explicitly focus on approximating the human users model, which is central to this paper. 

\subsection{Human Model Approximation}

Accurate human models are crucial for effective human-AI interactions. In argumentation, several approaches have emerged. \citet{rienstra2013opponent} proposed a probabilistic opponent model for move selection based on perceived awareness. \citet{hadjinikolis2013opponent} explored dialogue history analysis to predict opponent arguments. \citet{hadoux2015optimization} introduced probabilistic finite state machines and partially observable Markov decision processes for modeling dialogue progression under uncertainty.

In other domains, various approaches to human model approximation exist. Deep learning has been used to simulate and predict human behavior from large datasets \citep{HAMRICK20198,lake2017building}. Game-theoretic models reveal how agents' mental states affect choices and strategies in competitive scenarios \citep{yoshida2008game,camerer2011behavioral}. Planning formalisms have been utilized to learn human models in human-AI interaction settings \cite{sreedharan2018handling,ijcai2018-671,black2014automated}. The problem of learning human preferences has also been extensively studied, particularly in recommendation systems \cite{furnkranz2010preference}. Preferences are often elicited via ranking or comparisons \cite{ailon2012active,wirth2017survey}, or reinforcement learning paradigms \cite{wilson2012bayesian,biyik2022aprel}.

 

Most relevant to our work are those by \citet{hunter2013probabilistic, hunter2015modelling, hunter2016persuasion}, which present methods for representing and updating human beliefs through probability distributions during persuasion dialogues. While they provided essential theoretical groundwork, our approach extend them by incorporating insights from prospect theory and introducing personalized modeling capabilities that account for individual differences in probability assessment.
 

As our approach is specifically designed for approximating human models in argumentation-based dialogues, we compare it to the most relevant work in this space, namely the work by \citet{hunter2015modelling,hunter2016persuasion}. We do not compare it against non-argumentation approaches, as they lack the specific structures and mechanisms necessary for handling structured arguments and belief updates in dialogue settings. Our focus on argumentative reasoning and uncertainty in dialogues requires specialized techniques that these general approaches do not provide.

\section{Background}
We will use classical propositional logic to describe aspects of the world. Consider a finite (propositional) language $\fml{L}$ that utilizes the classical entailment relation, represented by $\models$. The set of \textit{models} (i.e.,~possible words) of $\fml{L}$ is denoted by $\fml{M}$, where each model $m_i \in \fml{M}$ is an assignment of true or false to the formulae of $\fml{L}$ defined in the usual way for classical logic. For $\phi \in \fml{L}$, let~$\texttt{Mod}(\phi)=\{m_i \in \fml{M} \: | \: m_i \models \phi \}$ denote the set of all models of $\phi$.  

Building on a propositional language $\fml{L}$, we model the uncertainty of arbitrary formulae using a \textit{probability distribution} over the models $\fml{M}$ of $\fml{L}$:

\begin{definition}[Probability Distribution]
Let $\fml{M}$ be the set of models of the language $\fml{L}$. A probability distribution $P$ on $\fml{M}$ is a function $P: \fml{M} \mapsto [0,1]$ such that 
$\underset{m \in \fml{M}}{\sum}P(m) = 1$.
\end{definition}

In essence, the probability distribution allows an agent to create a \textit{ranking} between possible words with respect to how likely they are to be true. This then allows the agent to compute the \textit{probability of a formula} as follows:

\begin{definition}[Probability of Formula]
\label{def:prob-formula}
Let $\fml{M}$ be the set of models of language $\fml{L}$ and $P$ a probability distribution over $\fml{M}$. The probability of formula $\phi \in \fml{L}$ is $P(\phi) = \underset{m \models \phi}{\sum}P(m)$.
\end{definition}

\xhdr{Argumentation-based Dialogues:}
In an argumentation-based dialogue, agents take turns exchanging arguments that prove (or disprove) specific claims, where the structure and relationships between these arguments are governed by the underlying argumentation semantics \cite{black2021argumentation}. In this paper, we consider the semantics of structured (deductive) argumentation \cite{besnard2014constructing}, where each argument is constructed using formulae from language $\fml{L}$. Formally,

\begin{definition}[Argument]\label{definition:argument}
Let $\fml{L}$ be the language and $\phi \in \fml{L}$ a formula. Then,
\squishlist
\item \textbf{Argument:} $A= \langle \Phi, \phi \rangle$ is an argument for $\phi$ iff: (i) $\Phi \subseteq \fml{L}$; (ii) $\Phi \models \phi$; (iii) $\Phi \not \models \perp$; and (iv) $\nexists \Phi^{\prime} \subset \Phi$ s.t. $\Phi^{\prime} \models \phi$. We refer to $\phi$ as the \emph{claim} of the argument, and $\Phi$ as the \emph{premise} of the argument.

\item \textbf{Attack Relation:} 
 An argument $A'=\langle\Psi, \psi \rangle$ \emph{attacks} argument $A= \langle \Phi, \phi \rangle$ and vice versa (i.e., $A$ attacks $A'$) iff $\Phi \cup \Psi \models \perp$, where $\bot$ denotes falsity.
\squishend
\end{definition}

The attack relation captures conflicts between arguments, which is essential for modeling disagreements.

\begin{example}
Let $\mathcal{L}$ be the language consisting of variables $\{a, b, c\}$. Then, $A_1 = \langle \{b, b \rightarrow a\}, a \rangle$ and $A_2 = \langle \{\neg c, \neg c \rightarrow \neg a\}, \neg a \rangle$ are two arguments for $a$ and $\neg a$, respectively. Note that $A_1$ attacks $A_2$ and vice versa, as $\{b, b \rightarrow a, \neg c, \neg c \rightarrow \neg a\} \models \bot$.
\end{example}

In real-world scenarios, arguments often come with some degree of uncertainty. We can capture this uncertainty with a probability distribution over the models $\fml{M}$ of $\fml{L}$, and then use it to compute the probability for any argument $A= \langle \Phi, \phi \rangle$ using Definition~\ref{def:prob-formula}, i.e., $P(A) = \sum_{m \models A} P(m)$, where $m \models A$ is a shorthand notation to mean that the premise $\Phi$ of $A$ is true in $m$.

Now, in this paper, we are mainly interested in argumentation-based \textit{dialogue traces} between two agents, i.e., finite sequences of arguments that attack each other:

\begin{definition}[Dialogue~Trace]\label{def: dialogue_traces}
Let $\Delta$ be an argumentation-based dialogue between agents $\alpha$ and $\eta$. A dialogue trace from $\Delta$ is defined as $\fml{T} = \langle (A_1, x_1)^{t_1}, (A_2, x_2)^{t_2}, \ldots, (A_n, x_n)^{t_n} \rangle$, where each $(A_i, x_i)^{t_i}$ denotes the argument put forward by agent $x_i \in \{\alpha, \eta \}$ at timestep $t_i$. 
\end{definition}

\begin{figure}[!t]
    \centering
    \includegraphics[width=0.35\linewidth]{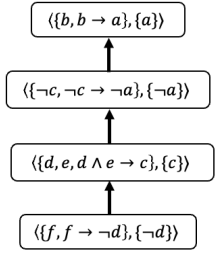}
    \caption{Example dialogue trace $\fml{T}$ between two agents.}
    \label{fig:dialogue-trace}
\end{figure}

A dialogue trace $\fml{T}$ can also be represented as a tree with $n$ nodes and $n-1$ edges, where each node $i$ represents the argument expressed at timestep $t_i$, and there is a directed edge from node $j$ to node $i$ iff argument $A_j$ attacks argument $A_i$, where $1 \leq i < j \leq n$. 
Note that repeating arguments within the same dialogue trace are not allowed to avoid infinite loops. Figure~\ref{fig:dialogue-trace} shows an example of the dialogue trace.

\section{Approximating Human Models}

We now introduce our framework that enables an agent to progressively update its approximation of the human model through argumentation-based dialogue traces.

\xhdr{Problem Setting and Assumptions:} Our setting consists of an AI agent (denoted $\alpha$) interacting with a human user (denoted $\eta$) via an argumentation-based dialogue. We make the following key assumptions:

\squishlist
    \item \textbf{Shared Domain Language}: Both $\alpha$ and $\eta$ have access to and communicate in the same language $\mathcal{L}$ using a shared vocabulary of atomic variables. This allows them to construct domain-specific formulae.      
        
    \item \textbf{Probabilistic Human Model}: The human model is represented as a probability distribution $P_h^{t_i}$ over the possible models $\mathcal{M}$ of $\mathcal{L}$ at each timestep $t_i$. Initially, we assume a uniform distribution: $P_h^{t_0}(m) = \frac{1}{|\mathcal{M}|}$ for all $m \in \mathcal{M}$.
 
    \item \textbf{Dialogue Traces}: We have access to (finite) dialogue traces $\mathcal{T}$ produced by argumentation-based dialogues between $\alpha$ and $\eta$ \cite{vasileiou_DR}.
    
\squishend

In real-world argumentation, arguments often come with some degree of uncertainty. To capture this, we associate a probability $p(A_i)$ with each argument $A_i$ in the dialogue trace. It is crucial to note that these probabilities represent uncertainty from the perspective of the human user, that is, how likely the human thinks that the argument is true.\footnote{These probabilities can arise from various sources, such as incomplete or imprecise knowledge, subjective interpretations, or lack of confidence in the reasoning process. Accounting for these uncertainties is crucial for developing a more realistic and nuanced model of human reasoning in argumentative contexts.
}

\subsection{Updating the Human Model}

Given a dialogue trace $\mathcal{T}$, we employ a Bayesian approach to update the agent's probability distribution $P_h$ over possible human models. At each timestep $t_i$, when an argument $A_i$ is presented, we perform the following update:
\begin{equation}\label{eq:belief_update}
P_h^{t_i}(m) = \left\{
\begin{aligned}
& \frac{P_h^{t_{i-1}}(m)}{\sum_{m \models A_i} P_h^{t_{i-1}}(m)} \cdot p(A_i) & \text{if } m \models A_i  \\
& \frac{P_h^{t_{i-1}}(m)}{\sum_{m \not\models A_i} P_h^{t_{i-1}}(m)} \cdot (1-p(A_i)) & \text{if } m \not\models A_i
\end{aligned}
\right.
\end{equation}
This update mechanism increases the probability of human models that are consistent with the presented argument, weighted by the argument's associated probability $p(A_i)$. Models that are inconsistent with the argument have their probabilities decreased accordingly.

\xhdr{A More Personalized Approach to Uncertainty Estimation:}
While the Bayesian update approach provides a solid foundation for estimating the human model, it does not account for the subjective nature of how humans perceive and think about uncertainty. To address this, we introduce a more personalized approach based on prospect theory~\cite{kahneman1979prospect}, which allows us to capture individual differences in how humans evaluate probabilities in argumentative contexts.\footnote{According to prospect theory, humans tend to overweight small probabilities and underweight large probabilities.}

\begin{figure}[t]
\centering
\includegraphics[width=0.45\textwidth]{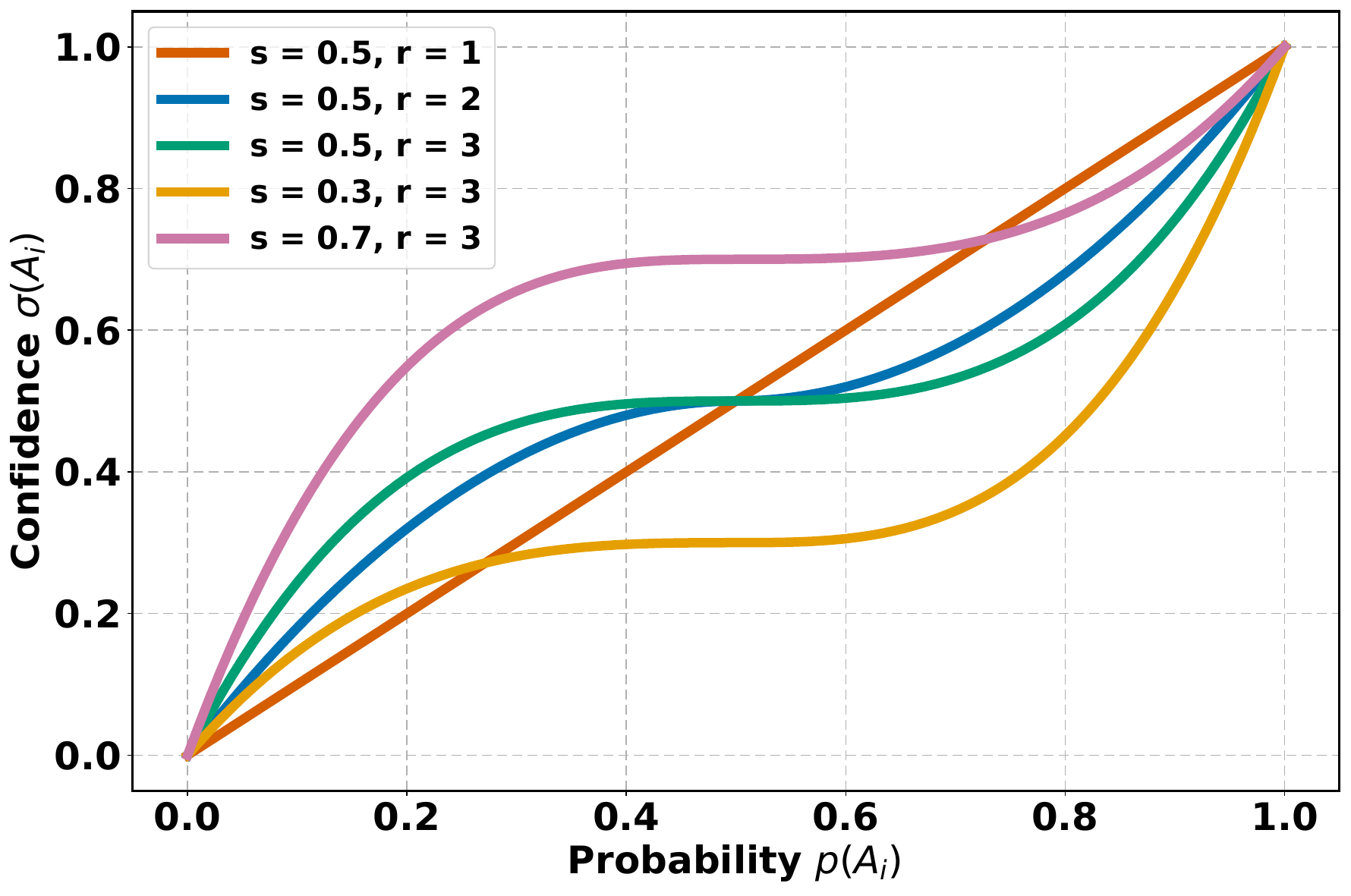}
\caption{Probability weighting function with value pairs $(s, r)$ given as $\{(0.5, 1), (0.5, 2),(0.5, 3), (0.3, 3), (0.7, 3)\}$. Lower values of $s$ imply that the average probability reflects a lower level of human confidence in the agent's argument, whereas higher values of $r$ suggest excessive distortion, either through overweighting or underweighting of the probability. Note that the relationship between confidence and probability is linear when $s = 0.5$ and $r = 1$.}
\label{fig:weighting_function} 
\end{figure}

We propose the following probability weighting function to model the relationship between ``actual'' probabilities $p(A_i)$ and subjective probability perceptions, referred to as \textit{confidence values} $\sigma(A_i)$:
\begin{equation}
\sigma(A_i)= \begin{cases}
s+(1-s) \cdot(2 \cdot p(A_i)- 1)^r & \text { if } p(A_i) >0.5 \\ 
s - s \cdot (1 - 2\cdot p(A_i))^r & \text { if } p(A_i) \leq 0.5
\end{cases}
\label{eq:confidence_to_prob}
\end{equation}
where parameter $s \in (0,1)$ determines the value of $\sigma(A_i)$ when $p(A_i) = 0.5$, regardless of the value of $r$, and parameter $r \in [1, \infty)$ controls the degree of this nonlinear distortion. The relationship between $s$ and $r$ is shown in Figure~\ref{fig:weighting_function}.

In practice, we often need to infer the probability $p(A_i)$ from the observed subjective probability $\sigma(A_i)$. To do this, we invert Equation~\ref{eq:confidence_to_prob}:
\begin{equation}
p(A_i)= \begin{cases}
\frac{1}{2} - \frac{1}{2} \cdot (\frac{s - \sigma(A_i)}{s})^{\frac{1}{r}}  & \text { if } \sigma(A_i) \leq s \\ 
\frac{1}{2} + \frac{1}{2} \cdot (\frac{\sigma(A_i) - s)}{1 - s})^{\frac{1}{r}} & \text { if } \sigma(A_i) > s
\end{cases}
\label{eq:confidence_to_prob_inverse}
\end{equation}

\begin{example}
Consider the dialogue trace shown in Figure~\ref{fig:dialogue-trace}. At timestep $t_1$, the agent asserts the argument $A_1 = \langle \{b, b \rightarrow a \}, \{a\} \rangle$. The human assigns a confidence value of $\sigma(A_1) = 0.6$ to this argument. Assuming $s = 0.5$ and $r = 1.5$, the actual probability of $A_1$ is computed using Equation~\ref{eq:confidence_to_prob_inverse}:
$$p(A_1)= \frac{1}{2} + \frac{1}{2} \cdot (\frac{0.6 - 0.5}{ 1 - 0.5})^{\frac{1}{1.5}} \approx 0.67$$

Suppose there are eight possible models, $\mathcal{M} = \{m_1, m_2, \ldots , m_8\}$, with a uniform prior distribution $P_h^{t_0}(m_1) = \ldots = P_h^{t_0}(m_8) = 0.125$. Let $m_1$ and $m_2$ be the models that entail argument $A_1$, i.e., $m_1, m_2 \models A_1$. Applying the update mechanism from Equation~\ref{eq:belief_update}, we get:
\begin{equation*}
\begin{aligned}
 P_h^{t_1}(m_1) & =  P_h^{t_1}(m_2) = \frac{0.125}{0.125 + 0.125} \cdot 0.67 = 0.335 \\
 P_h^{t_1}(m_3) & = P_h^{t_1}(m_4) = P_h^{t_1}(m_5) = P_h^{t_1}(m_6) = P_h^{t_1}(m_7) \\
 & = P_h^{t_1}(m_8) =\frac{0.125}{0.125 \cdot 6} \cdot 0.33 = 0.055 
\end{aligned}
\end{equation*}

After this update, the models $m_1$ and $m_2$ that are consistent with the agent's argument have a higher probability than the other six models, reflecting the human's moderate confidence in the argument.

At the next timestep $t_2$, the human presents the argument $A_2 = \langle \{\neg c, \neg c \rightarrow \neg a\}, \{\neg a \} \rangle$ with probability $p(A_2) = 0.9$. Let $m_3$ and $m_4$ be the models that entail argument $A_2$. Applying the update mechanism again, we get:
\begin{equation*}
    \begin{aligned}
    P_h^{t_2}(m_1) &= P_h^{t_2}(m_2) =\frac{0.335}{0.335\cdot 2 + 0.055 \cdot 4} \cdot 0.1 = 0.038 \\
    P_h^{t_2}(m_3) &= P_h^{t_2}(m_4) = \frac{0.055}{0.055+0.055} \cdot 0.9 = 0.45 \\
    P_h^{t_2}(m_5) &= P_h^{t_2}(m_6) = P_h^{t_2}(m_7) = P_h^{t_2}(m_8) \\
    &= \frac{0.055}{0.335\cdot 2 + 0.055 \cdot 4} \cdot 0.1 = 0.006 
    \end{aligned}
\end{equation*}

After this update, the models $m_3$ and $m_4$ that are consistent with the human's argument have a much higher probability than the models consistent with the agent's previous argument. The same process can be applied in the remaining two timesteps.
\end{example}

\xhdr{Personalized Parameter Learning:} To adapt the personalization parameters $s$ and $r$ to individual users, we can use a data-driven approach based on dialogue traces and user-provided model rankings. This method aims to find the optimal parameters that maximize the correlation between our computed model rankings and the ground truth rankings provided by users.

The approach involves collecting two types of data from users: Dialogue traces and model rankings. For a given pair of parameters $(s,r)$, we compute a ranking over models using the probability weighting function and belief update mechanism described earlier. We then evaluate the fit of parameters $(s,r)$ by computing the correlation between our computed ranking and the user-provided ground truth ranking. The optimal parameters $(s^*, r^*)$ are then determined by maximizing the correlation:
\begin{equation}
    (s^*, r^*) = \argmax_{(s,r)} \rho(s,r)
    \label{eq:optimize_correlation}
\end{equation}
where $\rho(s,r)$ denotes the correlation between computed and ground truth rankings.

The specific implementation details, including how we split the data for learning and evaluation, is discussed in the empirical evaluation section.

\section{Empirical Evaluations}

We now evaluate the effectiveness of our approach across two dimensions: (1) Its ability to personalize and optimize the probability weighting function parameters; and (2) Its performance in approximating human models and estimating argument probabilities compared to existing methods. To collect data, we conducted the following human-subject study.

\subsection{Human-Subject Study Description}

We simulated a scenario where participants interacted with an AI assistant named \textit{Blitzcrank} to evaluate the suitability of a fictional venue, \textit{Luminara Gardens}, for a company team-building event. This scenario was chosen to provide a concrete context for argumentation while being accessible to a general participant pool.

The study consisted of a series of interaction rounds (maximum 5) between each participant and Blitzcrank. Each round followed this structure:
\squishlist
    \item Blitzcrank presented an argument about Luminara Gardens' suitability.
    \item Participants rated their confidence in Blitzcrank's argument on a five-point scale: Very low (0.1), low (0.3), average (0.5), high (0.7), or very high (0.9).
    \item Participants selected and presented a counterargument to Blitzcrank from a set of three options, each associated with a confidence level.
    \item Participants ranked four different perspectives (i.e., models) on Luminara Gardens' suitability.
\squishend

The dialogue continued for up to five rounds, with the option to end earlier if agreement was reached.

\xhdr{Data Collection:} We recruited 200 participants via the Prolific platform \cite{palan2018prolific}, ensuring a diverse sample.\footnote{Ethics approval was obtained through our university's IRB.} Participants were required to be fluent in English and were compensated USD 4.00 for their time. After applying attention checks and coherence filters, we retained data from 184 participants for analysis. For each participant~$i$, we collected:
\squishlist
    \item Dialogue traces \small$\mathcal{T}_i = \langle (\!A_1, x_1, \sigma_1)^{t_1}\!, \!\ldots\!, (\!A_{n_i}, x_{n_i}, \sigma_{n_i})^{t_{n_i}}\rangle$\normalsize,
    where $n_i \in \{8,10\}$ is the number of completed interactions, $x_j \in \{\text{Blitzcrank}, \text{Participant}\}$, and $\sigma_j$ is the participant's confidence value on argument $A_j$. 
    
    \item Model rankings $M_i^t = \langle m_1^t, m_2^t, m_3^t, m_4^t \rangle$ after each round $t$, where each round consists of two interactions (e.g., two exchanged arguments).
    
    \item Final argument rankings $R_i = \langle a_1, a_2, \ldots, a_m \rangle$, where $m$ is the total number of arguments presented.
    
    \item Post-study questionnaire responses assessing satisfaction and interaction quality.
\squishend

\subsection{Experiment 1: Learning Optimal Personalization Parameters}

Our first experiment aimed to learn the optimal values for the personalization parameters $s$ and $r$ in our probability weighting function (Equations \ref{eq:confidence_to_prob} and \ref{eq:confidence_to_prob_inverse}). This data-driven approach uses dialogue traces and user-provided model rankings to maximize the correlation between our computed model rankings and the ground truth rankings provided by the participants.

\xhdr{Methodology:}
For each participant $i$ with $n_i$ interactions, we performed the following steps: First, we iterated over each $k \in \{1, 2, 3 \}$ rounds of interactions. Then, for each $k$, we learn the optimal $(s^{k*}_{i}, r^{k*}_{i})$ for participant $i$ by:
\begin{equation}
    (s^{k*}_{i}, r^{k*}_{i}) = \argmax_{(s',r')} \sum_{t=1}^{k} \rho(M_i^t, \hat{M}_i^t(s',r'))
\label{eq:personalize_parameters}
\end{equation}
where $\rho$ is Spearman's rank correlation coefficient~\cite{spearmancorr}, $M_i^t$ is the participant's model ranking at $t$, and $\hat{M}_i^t(s',r')$ is the computed ranking using parameters $(s',r')$, where $s' \in \{0.1,0.2,\ldots,0.9\}$ and $r' \in \{1,2,\ldots,8\}$. Specifically, to compute $\hat{M}_i^t(s',r')$, we:
\squishlist
    \item Used Equation \ref{eq:confidence_to_prob_inverse} to transform confidence values $\sigma_j$ in the dialogue trace to probabilities.
    \item Applied the belief update mechanism (Equation \ref{eq:belief_update}) to compute the distribution $P_h^t(m)$ over models.
    \item Ranked the models based on their probabilities in $P_h^t(m)$.
\squishend

\noindent We then evaluated the learned optimal values $(s^{k*}_{i}, r^{k*}_{i})$ for each participant $i$ in a future round $k' > k$:
\begin{equation}
\label{eq:eq6}
    \rho_i^{k'} = \rho(M_i^{k'}, \hat{M}_i^{k'}(s^{k*}_i , r^{k*}_{i}))
\end{equation}

This approach allows us to assess how well the learned parameters generalize to new, unseen interactions in round $k'$. By varying $k$, we can analyze how the amount of training data affects the model's performance.

\begin{figure}[t]
\centering
\includegraphics[width=0.45\textwidth]{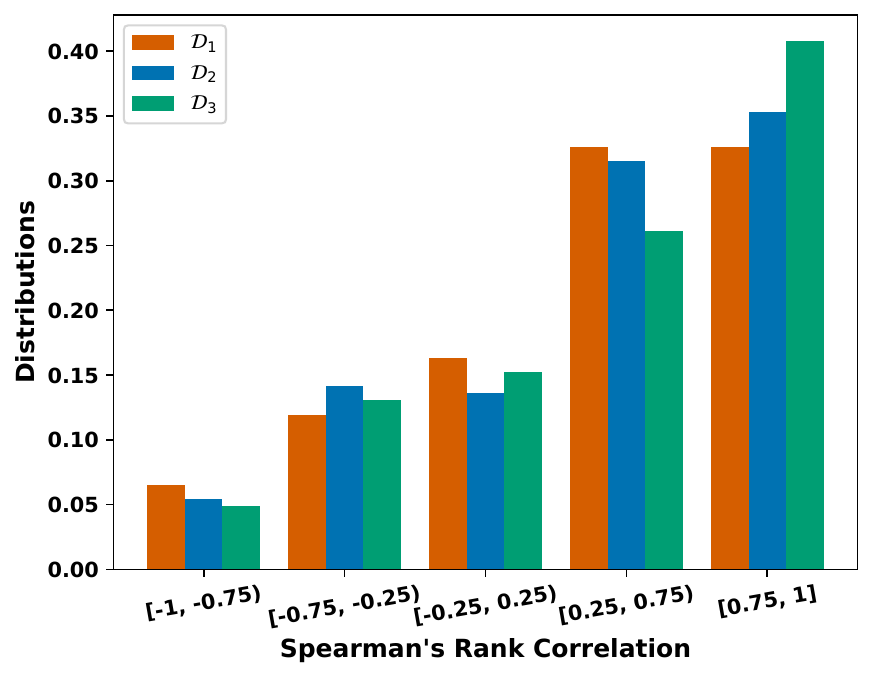}
\caption{Comparisons of Spearman's rank correlation distributions in round four derived from the first $k$ rounds of interaction data $\mathcal{D}_k$.}
\label{fig:spearman_rank_correlation_prediction} 
\end{figure}

\xhdr{Evaluation Metrics:} We use Spearman's Rank Correlation $\rho$ between computed and ground truth rankings, and Paired Student's $t$-tests to assess the statistical significance between the evaluated methods.

\begin{table}[t]
\centering 
\small
\begin{tabular}{cccc}
\toprule[2pt]
\diagbox{$X$}{$Y$}& $\mathcal{D}_1$ & $\mathcal{D}_2$ & $\mathcal{D}_3$\\ \midrule[2pt] 
$\mathcal{D}_1$ & --    & $0.925$  & $1$ \\ [0.2ex]
$\mathcal{D}_2$ &  $0.075$  & --  & $0.985$   \\ [0.2ex]
$\mathcal{D}_3$ &   $4.43 \times 10^{-4}$ & $0.015$   & --  \\ \bottomrule[2pt]
\end{tabular}
\caption{The $p$-values from Student's $t$-tests assessing the hypothesis that $X$ outperforms $Y$ in Experiment 1.}
\label{table:p-value_personalization}
\end{table}

\xhdr{Results:}
Figure~\ref{fig:spearman_rank_correlation_prediction} shows the distribution of Spearman's rank correlation coefficients in round four (i.e.,~$k' = 4$ in Equation~\ref{eq:eq6}) using the optimized parameters from the first $k$ interaction data $\mathcal{D}_{k}$. We applied the learned parameters in round four since the minimum number of rounds for all participants was four.
A high positive correlation indicates method effectiveness, with particular attention given to distributions above 0.75. As the value of $k$ increases, our approach can better approximate the human model for the fourth round by leveraging more data from previous rounds, enhancing parameter personalization for each participant. In this way, the more data we use to learn personalized parameters, the more accurately the human model is approximated. 

To further assess the personalization component of Persona with different values of $k$, we conducted paired Student's $t$-tests. Table~\ref{table:p-value_personalization} presents the $p$-values, evaluating the hypothesis that $X$ (rows of the table) outperforms $Y$ (columns). The results show that $\mathcal{D}_3$ statistically significantly outperforms $\mathcal{D}_1$ and $\mathcal{D}_2$, with $p$-values smaller than $0.05$. While $\mathcal{D}_2$ does outperform $\mathcal{D}_1$, there is no statistical significance, as the $p$-value is greater than $0.05$.

\subsection{Experiment 2: Comparative Evaluation}

The goal of our second experiment is two-fold: (1)~To evaluate the effectiveness of our approach on approximating human models; and (2)~To evaluate the effectiveness of our approach on estimating the human beliefs of arguments. We used the same evaluation metrics as in Experiment 1.

\subsubsection{Experiment 2.1: Human Model Approximation\\}

In this experiment, we evaluated the efficacy of our personalized approach, referred to as \textit{Persona} in subsequent figures, in approximating human models. We compared our method against the following baselines:\footnote{Please refer to the supplement in our GitHub repository for details about the baselines.}

\squishlist
    \item \textit{Generic}: Instead of personalizing parameters for each participant, we learned the same $(s, r)$ for each participant in the first $k$ rounds, i.e., Equation \ref{eq:personalize_parameters} can be modified as:
    \begin{equation}
    (s^{k*}, r^{k*}) = \argmax_{(s',r')} \sum_{i}\sum_{t=1}^{k} \rho(M_i^t, \hat{M}_i^t(s',r'))
    \end{equation}
    This serves as an ablation study for Persona.

    \item \textit{SBU}: The simple Bayesian update we proposed in Equation~\ref{eq:belief_update}. This serves as an ablation study for Persona as well.
    
    \item \textit{$H\!M_1$}: An argumentation-based method for updating probability distributions of human models based on argument graphs \cite{hunter2015modelling}.
    \item \textit{$H\!M_2$}: An enhanced version of Hunter's $H\!M_1$ that utilizes the argument structure for updating the distribution \cite{hunter2015modelling}. 
\squishend

\begin{figure}[t]
\centering
\includegraphics[width=0.45\textwidth]{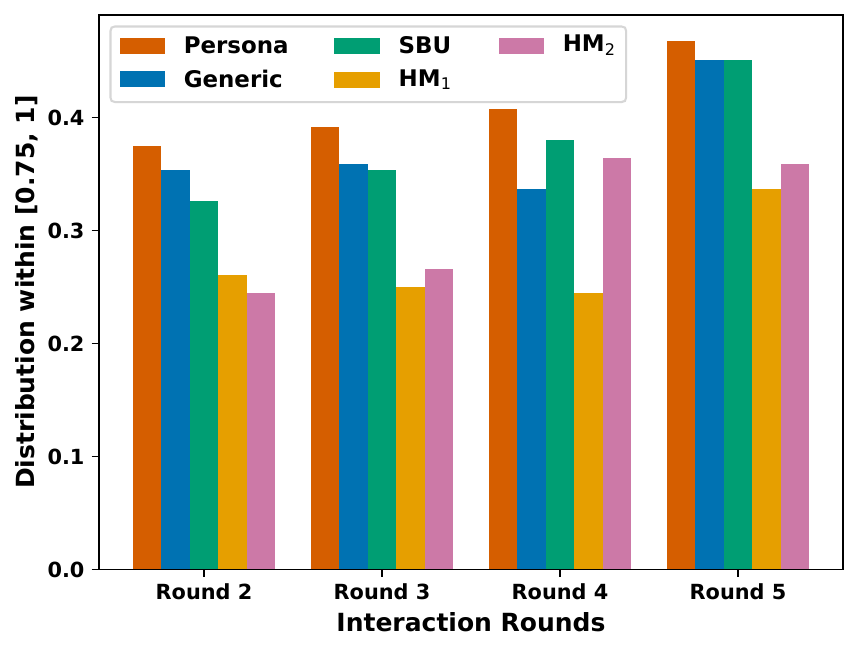}
\caption{Comparisons of Spearman's rank correlation distributions in model estimation within $[0.75, 1]$ in round $k$ $(k = 2,3,4,5)$ of human model rankings where parameters are learned from the previous $k-1$ rounds. Note that for participants with only four interactions, the results for Round 5 are identical to those of Round 4.}
\label{fig:spearman_rank_correlation_model_comparison} 
\end{figure}

\begin{table}[t]
\centering
\resizebox{1.\columnwidth}{!}{ 
\begin{tabular}{ccccc}
\toprule[2pt]
\diagbox{$X$}{Rounds}& Round 2 & Round 3 & Round 4 & Round 5\\ \midrule[2pt] 
\textit{Generic}   & $0.043$    & $0.004$  & $0.010$ &  $0.640$ \\ [0.2ex]
\textit{SBU}    &  $0.047$  & $0.003$  & $0.047$ & $0.426$  \\ [0.2ex]
\textit{$H\!M_1$}     & $2.408 \times 10^{-6}$   & $3.255 \times 10^{-4}$  & $3.760 \times 10^{-5}$  & $0.006$ \\ [0.2ex]
\textit{$H\!M_2$}   &  $5.730 \times 10^{-5}$   & $0.002$  & $0.006$ & $0.001$  \\ \bottomrule[2pt]
\end{tabular}
}
\caption{The $p$-values from Student's $t$-tests assessing the hypothesis that Persona outperforms $X$ in Experiment 2.1.}
\label{table:p-value_models}
\end{table}

\xhdr{Results:} We compared the Spearman’s rank correlation distributions in round $k = \{2,3,4,5\}$ of human model rankings where parameters are learned from the previous $k-1$ rounds among Persona and its ablations and the two baselines. 
Figure~\ref{fig:spearman_rank_correlation_model_comparison} displays the distribution of Spearman's rank correlation coefficients for interval $[0.75, 1]$ for human models.\footnote{We omit Spearman's rank correlation coefficients in other intervals, but we describe them in the supplement.} We observed that Persona performed better than all the other methods in all rounds.
Compared to \textit{Generic} and \textit{SBU}, the results demonstrate that incorporating both personalization and the weighting function increases the accuracy of model approximation. 
Notably, Persona significantly outperformed $H\!M_1$ and across all rounds, and $H\!M_2$ across rounds $2$, $3$, and $5$. Interestingly, Hunter's $H\!M_2$ has close results to Persona in round $4$ due to the randomness of the method during the ranking procedure.

We also conducted paired Student's $t$-tests among various methods, where Table \ref{table:p-value_models} presents the $p$-values evaluating the hypothesis that Persona outperforms method $X$ in human model approximation across different rounds.
The results demonstrate that Persona statistically significantly outperforms all the other methods in almost all rounds. These findings underscore Persona's capacity to effectively utilize existing data to learn personalized parameters for each participant, thereby enhancing the accuracy of human model estimation in the subsequent round compared to state-of-the-art baselines and the ablation variants.

\subsubsection{Experiment 2.2: Human Argument Belief Estimation\\}

In this experiment, we evaluated the efficacy of Persona in estimating argument beliefs. For baselines, we used the state-of-the-art method for learning probability distributions of arguments by \citet{hunter2016persuasion}, referred to as $H\!A$, as well as the ablation variants used in Experiment 2.1. 
Unlike in Experiment 2.1, we only use data from a subset of individuals from the human-subject study. Specifically, we omitted data of individuals whose dialogue traces ended with the agent because, for these individuals, there were only \emph{two} relevant arguments that they needed to rank. In contrast, the other group of individuals who ended their conversations had \emph{four} arguments that they needed to rank. Two arguments are, in our opinion, too few for meaningful comparisons. 

\begin{figure}[t]
\centering
\includegraphics[width=0.45\textwidth]{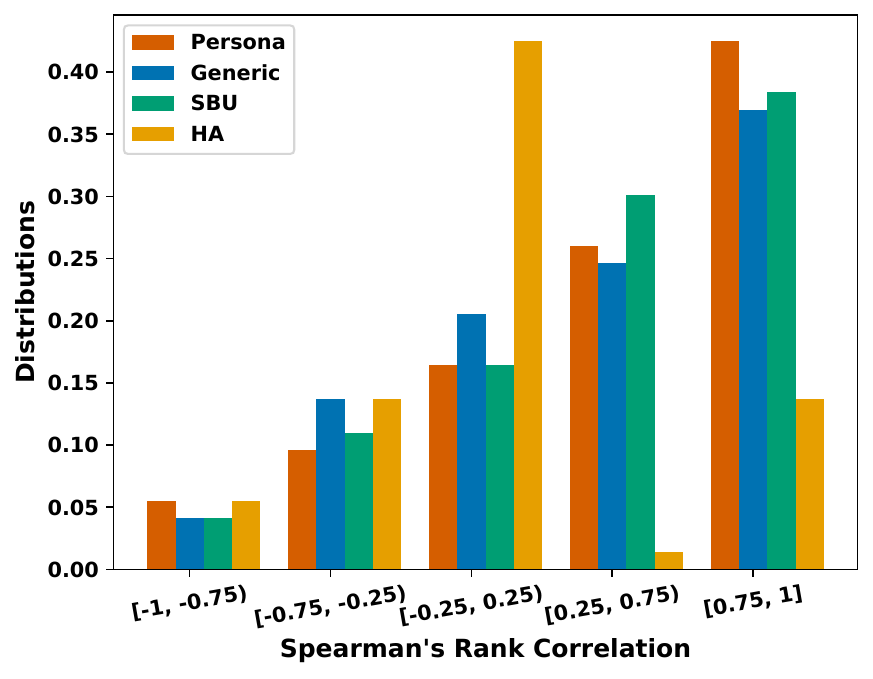}
\caption{Comparisons of Spearman's rank correlation distributions for argument beliefs across different methods.}
\label{fig:spearman_rank_correlation_arguments} 
\end{figure}

\xhdr{Results:} Figure \ref{fig:spearman_rank_correlation_arguments} displays the distribution of Spearman's rank correlation coefficients for argument beliefs. 
Persona applies the personalized values of $s$ and $r$ during the first $n_i-1$ rounds for each participant $i$ in argument belief estimation, while \textit{Generic} uses fixed values of $s$ and $r$ learned from the first three rounds across all participants. Focusing on the range $[0.75, 1]$ of high correlation, our results show that Persona outperforms its ablation variants, demonstrating the benefits of learning and personalization in improving argument belief estimations. Additionally, it surpasses the $H\!A$ method as well.

To better understand the statistical significance of this observation, we also conducted paired Student's $t$-tests. Table \ref{table:p-value_arguments} presents the $p$-values evaluating the hypothesis that method $X$ (rows) outperforms method $Y$ (columns) in argument belief approximation. 
Surprisingly, there is no statistical difference between Persona and \textit{SBU}, as the $p$-value is significantly larger than 0.05. The reason is that while Persona does better than \textit{SBU} in the high correlation range, \textit{SBU} does better in the other ranges.
However, the improvement of Persona over \textit{Generic} and $H\!A$ are statistically significant, with $p$-values smaller than 0.05. Additionally, especially noteworthy are the extremely small $p$-values for Persona and the two ablation variants over $H\!A$, highlighting the strength of our framework against the state of the art.

\begin{table}[t]
\centering 
\small
\begin{tabular}{ccccc}
\toprule[2pt]
\diagbox{$X$}{$Y$}& \textit{Persona} & \textit{Generic} & \textit{SBU} & $H\!A$\\ \midrule[2pt] 
\textit{Persona} & --    & $0.002$  & $0.569$ & $2.105 \times 10^{-11}$\\ [0.2ex]
\textit{Generic} &  $0.998$  & --  & $0.999$ & $5.674 \times 10^{-8}$    \\ [0.2ex]
\textit{SBU} &   $0.431$ & $0.001$   & -- & $1.410 \times 10^{-11}$ \\ [0.2ex]
$H\!A$ &      $1$  & $1$ & $1$ & -- \\ \bottomrule[2pt]
\end{tabular}
\caption{The $p$-values from Student's $t$-tests assessing the hypothesis that $X$ outperforms $Y$ in Experiment 2.2.}
\label{table:p-value_arguments}
\end{table}

\subsection{Computational Results}

We implemented Persona and evaluated its performance on a MacBook Pro with a 2.2 GHz Quad-Core Intel Core i7 processor and 16GB of RAM.  Persona took approximately 0.6 seconds to compute probabilities for each $s$ and $r$ pair of hyperparameter values per participant. In comparison, methods $H\!M_1$ and $H\!M_2$ took around 9 seconds and 41 seconds per participant, respectively, whereas $H\!A$ required just 0.003 seconds to compute argument beliefs. These runtimes indicate that all approaches, particularly Persona, are suitable for real-time evaluations in practical applications. However, it is important to note that additional time would be required for translating between natural language and logic, which is an area we plan to address in future work.

\section{Conclusions and Future Work}
In this paper, we introduced Persona, a novel framework for personalizing human model approximations in argumentation-based dialogues. Persona combines a Bayesian belief update mechanism that refines probability distributions over potential human models during dialogues with a prospect theory-inspired probability weighting function. This combination allows for the incorporation of uncertainty estimates for both agent and human arguments while capturing individual differences in how humans evaluate probabilities in argumentative contexts.

Through a comprehensive human-subject study involving 184 participants, we demonstrated the effectiveness of Persona in both model approximation and argument belief estimation. Our empirical evaluations showed that Persona significantly outperforms state-of-the-art methods in terms of Spearman's rank correlation and statistical significance tests. Furthermore, our computational results indicate that Persona is suitable for practical applications, with competitive runtime performance compared to existing methods.

For future work, we plan to investigate how these learned human models can be used to generate more persuasive arguments 
as well as apply them to other applications, including automated planning~\cite{chakraborti2017plan,sreedharan2020emerging,vasileioulogic,vasileiouplease} and scheduling~\cite{vcyras2019argumentation,agrawal2020using,pozanco2022explaining,vasileioua2023lasp}.

\section*{Acknowledgments}
This research is partially supported by the National Science Foundation under award 2232055 and by J.P. Morgan AI Research. The views and conclusions contained in this document are those of the authors and should not be interpreted as representing the official policies, either expressed or implied, of the sponsoring organizations, agencies, or the United States government.

\bibliography{aaai25}

\newpage
\section{Details of Empirical Evaluations and Additional Results}

We provide the details of baselines and additional evaluation results below.

\subsection{Baselines in Comparative Evaluation}

\squishlist
\item \textit{$H\!M_1$}: An argumentation-based method for updating probability distributions of human models based on argument graphs \cite{hunter2015modelling}.
Inspired by the redistribution function, we apply this concept to our model distribution update.  

At each time step $t_i$, when an argument $A_i$ is presented by either the agent or the human, we perform the following naive update on the probability distribution: 
    \begin{equation}
	P^{t_i}_{h}(m)= 
	\begin{cases}
		P^{t_i - 1}_{h}(m)+  P^{t_i - 1}_{h}\left(h_{A_i}(m)\right) & \text { if } m \models A_i \\ 
		0 & \text { if } m \not \models A_i
	\end{cases}
	\label{eq:redistribution}
\end{equation}
where $h_{A_i}(m)=m \backslash\{\alpha\}$ and $\alpha$ is of the form $A_i$. 

\item \textit{$H\!M_2$}: An enhanced version of Hunter's $H\!M_1$ that utilizes the argument structure for updating the distribution \cite{hunter2015modelling}. Specifically, consider an argument graph $G$ where $\operatorname{Attacks}(G)$ represents the set of attack relations in $G$. For instance, if $A_1 = \langle \{a\}, \{a\} \rangle$ and $A_2 =  \langle \{b, b \rightarrow \neg a\}, \{\neg a\} \rangle$, $A_2$ is a counterargument of $A_1$, indicating that $(A_2, A_1) \in \operatorname{Attacks}(G)$. In this way, this method first applies Equation~\eqref{eq:redistribution} and then proceeds with the following update:
     \begin{equation}
     P^{t_i}_{h}(m)= 
     \begin{cases}
     P^{t_i}_{h}(m)+  P^{t_i }_{h}\left(h_{\Phi}(m)\right) & \text { if } m \models \Phi \\ 
     0 & \text { if } m \not \models \Phi
     \end{cases}
     \end{equation}
     where $\Phi = \{ \neg B \mid (B, A_i) \in \operatorname{Attacks}(G) \text{ or } (A_i, B) \in \operatorname{Attacks}(G)\}$.
    
\item \textit{$H\!A$}: A state-of-the-art method for learning probability distributions of arguments by \citet{hunter2016persuasion}. 
This baseline method updates the belief in each argument throughout the dialogue by considering the initial probability of each argument and the human's confidence in their arguments. 
Specifically, for each argument $A_i$, the final distribution is:
\begin{equation}
      P(A_i)\!\!= \begin{cases}
      0.2 & x_i = \alpha, \exists B \in \operatorname{Opp}(A_i), P(B)>0.5 \\ 
      0.2 & x_i = \eta, \exists B \in \operatorname{Pro}(A_i), P(B)>0.5 \\ 
      0.8 & x_i = \alpha, \forall B \in \operatorname{Opp}(A_i), P(B) \leq 0.5 \\ 
      \sigma_i & x_i = \eta, \forall B \in \operatorname{Pro}(A_i), P(B) \leq 0.5
      \end{cases}
      \end{equation}
where $\operatorname{Opp}(A_i)= \{A_{i+1} \mid \exists i, x_{i+1} = \eta\}$ and $\operatorname{Pro}(A_i) = \{A_j \mid \exists j, i<j, x_j = \alpha, (A_j, A_i) \in  \operatorname{Attacks}(G)\}$.
\squishend

\begin{example}
Suppose there are four possible models, $\mathcal{M} = \{m_1, m_2, m_3, m_4\}$, with a uniform prior distribution $P^{t_0}_h(m_1) = P^{t_0}_h(m_2) = P^{t_0}_h(m_3) = P^{t_0}_h(m_4) = 0.25$. At time step $t_1$, the agent asserts the argument $A_1 = \langle \{a\}, \{a\} \rangle$. At the next timestep $t_2$, the human presents the argument $A_2 =  \langle \{b, b \rightarrow \neg a\}, \{\neg a\} \rangle$ with confidence value $\sigma_2 = 0.6$. Applying the redistribution mechanisms in $H\!M_1$ and $H\!M_2$ in timesteps $t_1$ and $t_2$, respectively, the results are shown in Table \ref{table:example_of_hunter}. By applying $H\!A$ after timestep $t_2$, we have $P(A_1) = 0.2$ and $P(A_2) = 0.6$.
\begin{table}[t]
\centering \small
\begin{tabular}{ccccc}
\toprule[2pt]
& $m_1$ & $m_2$ & $m_3$ & $m_4$ \\ \midrule[2pt]
$a$                 & True  & True  & False & False \\ [1ex]
$b$                 & True  & False & True  & False \\ [1ex]
$P^{t_0}_{h}(m_i)$    & $0.25$   & $0.25$   & $0.25$   & $0.25$  \\ [1ex]
$P^{t_1}_{h}(m_i)$      & $0.5$   & $0.5$   &  $0$    &  $0$     \\
$P^{t_2}_{h}(m_i)$      & $0$   & $0$   &  $1$    &  $0$     \\
\bottomrule[2pt]
\end{tabular}
\caption{An example of the baseline methods.}
\label{table:example_of_hunter}
\end{table}
\end{example}

\subsection{Details of Human Model Approximation}

We compared the Spearman’s rank correlation distributions in round $k = \{2,3,4,5\}$ of human model rankings where parameters are learned from the previous $k-1$ rounds among Persona and its ablations and the two baselines. Table \ref{table:overall_model_comparison} displays the distribution of Spearman's rank correlation coefficients in each round, showing that Persona performed better than all the other methods in all rounds. Compared to \textit{Generic} and \textit{SBU}, the results demonstrate that incorporating both personalization and the weighting function increases the accuracy of model approximation. Notably, by considering the distribution within $[0.25, 1]$, Persona significantly outperformed $H\!M_1$ and $H\!M_2$ across all rounds.

We also performed paired Student's $t$-tests to compare various methods, with Table \ref{table:all_rounds_p_value} displaying the $p$-values that assess the hypothesis that method $X$ outperforms method $Y$ in human model approximation across different rounds. The results show that Persona consistently and statistically significantly outperforms all other methods in almost all rounds, with the exception of Round 5. The reason is that while Persona does better than \textit{Generic} and \textit{SBU} in the high correlation range, they both do better in the medium positive correlation range of $[0.25, 0.75)$ shown in Table \ref{table:round5_model_comparison}. However, the improvements of Persona and its ablation variants \textit{Generic} and \textit{SBU} over state-of-the-art baselines in every round are statistically significant, with $p$-values smaller than 0.05. 
These findings demonstrate Persona's ability to leverage existing data to personalize parameters, thereby enhancing human model estimation accuracy in subsequent rounds beyond state-of-the-art baselines and ablation variants. Notably, even the non-personalized ablation variants consistently outperform all baselines, further validating our approach.


\subsection{Post-study Results}
In our scenarios, participants were divided into Group A, who ended the conversation themselves, and Group B, where Blitzcrank ended the dialogue. Group A confirmed confidence levels across four rounds, while Group B did so over five rounds. The results in Table~\ref{table:t-test_confidence} provide compelling evidence that the confidence in the AI assistant increases, indicating that participants' confidence grows as the dialogue progresses and the assistant provides more relevant and persuasive arguments.

Finally, Table~\ref{table:comprehension_and_satisfaction} shows the post-study questionnaire responses further corroborate these findings, with participants reporting high levels of satisfaction with the interaction and the quality of Blitzcrank's arguments.

\begin{table}[H]
\centering 
\begin{subtable}[t]{\columnwidth}
\centering
\resizebox{1.\columnwidth}{!}{
\begin{tabular}{cccccc}
\toprule[2pt]
Round $2$ & \textit{Persona} & \textit{Generic} & \textit{SBU} & \textit{$H\!M_1$} & \textit{$H\!M_2$}\\ \midrule[2pt] 
$[-1, -0.75)$ & $0.059$    & $0.065$  & $0.065$ & $0.141$ & $0.152$ \\ [1ex]
$[-0.75, -0.25)$ &  $0.147$  & $0.141$  & $0.152$ & $0.185$ & $0.130$   \\ [1ex]
$[-0.25, 0.25)$  &   $0.233$ & $0.234$   & $0.207$ & $0.266$ & $0.234$  \\ [1ex]
$[0.25, 0.75)$ & $0.185$ & $0.207$ & $0.250$ & $0.147$ & $0.239$ \\ [1ex]
$[0.75, 1]$ & $0.375$ & $0.353$ & $0.326$ & $0.261$ & $0.245$ \\ \bottomrule[2pt]
\end{tabular}
}
\caption{Comparison of model estimation in Round 2.}
\label{table:round2_model_comparison}
\end{subtable}

\vspace{0.2em} 

\begin{subtable}[t]{\columnwidth}
\centering
\resizebox{1.\columnwidth}{!}{
\begin{tabular}{cccccc}
\toprule[2pt]
Round $3$ & \textit{Persona} & \textit{Generic} & \textit{SBU} & \textit{$H\!M_1$} & \textit{$H\!M_2$}\\ \midrule[2pt] 
$[-1, -0.75)$ & $0.059$    & $0.076$  & $0.076$ & $0.087$ & $0.081$ \\ [1ex]
$[-0.75, -0.25)$ &  $0.114$  & $0.120$  & $0.114$ & $0.179$ & $0.163$   \\ [1ex]
$[-0.25, 0.25)$  &   $0.217$ & $0.212$   & $0.234$ & $0.223$ & $0.245$  \\ [1ex]
$[0.25, 0.75)$ & $0.217$ & $0.234$ & $0.223$ & $0.261$ & $0.245$ \\ [1ex]
$[0.75, 1]$ & $0.391$ & $0.359$ & $0.353$ & $0.250$ & $0.266$ \\ \bottomrule[2pt]
\end{tabular}}
\caption{Comparison of model estimation in Round 3.}
\label{table:round3_model_comparison}
\end{subtable}

\vspace{0.2em}

\begin{subtable}[t]{\columnwidth}
\centering
\resizebox{1.\columnwidth}{!}{
\begin{tabular}{cccccc}
\toprule[2pt]
Round $4$ & \textit{Persona} & \textit{Generic} & \textit{SBU} & \textit{$H\!M_1$} & \textit{$H\!M_2$}\\ \midrule[2pt] 
$[-1, -0.75)$ & $0.049$    & $0.054$  & $0.054$ & $0.087$ & $0.071$ \\ [1ex]
$[-0.75, -0.25)$ &  $0.130$  & $0.130$  & $0.147$ & $0.158$ & $0.168$   \\ [1ex]
$[-0.25, 0.25)$  &   $0.152$ & $0.158$   & $0.152$ & $0.293$ & $0.185$  \\ [1ex]
$[0.25, 0.75)$ & $0.261$ & $0.321$ & $0.266$ & $0.217$ & $0.212$ \\ [1ex]
$[0.75, 1]$ & $0.408$ & $0.337$ & $0.380$ & $0.245$ & $0.364$ \\ \bottomrule[2pt]
\end{tabular}}
\caption{Comparison of model estimation in Round 4.}
\label{table:round4_model_comparison}
\end{subtable}

\vspace{0.2em}

\begin{subtable}[t]{\columnwidth}
\centering
\resizebox{1.\columnwidth}{!}{
\begin{tabular}{cccccc}
\toprule[2pt]
Round $5$ & \textit{Persona} & \textit{Generic} & \textit{SBU} & \textit{$H\!M_1$} & \textit{$H\!M_2$}\\ \midrule[2pt] 
$[-1, -0.75)$ & $0.060$    & $0.060$  & $0.060$ & $0.043$ & $0.065$ \\ [1ex]
$[-0.75, -0.25)$ &  $0.076$  & $0.081$  & $0.081$ & $0.136$ & $0.136$   \\ [1ex]
$[-0.25, 0.25)$  &   $0.147$ & $0.119$   & $0.130$ & $0.207$ & $0.185$  \\ [1ex]
$[0.25, 0.75)$ & $0.250$ & $0.288$ & $0.277$ & $0.277$ & $0.255$ \\ [1ex]
$[0.75, 1]$ & $0.467$ & $0.451$ & $0.451$ & $0.337$ & $0.359$ \\ \bottomrule[2pt]
\end{tabular}}
\caption{Comparison of model estimation in Round 5.}
\label{table:round5_model_comparison}
\end{subtable}

\caption{The distributions of Spearman’s rank correlation coefficients in model approximation in Round $k$ ($k = 2, 3, 4, 5$) of human model rankings where parameters are learned from the first $k-1$ rounds. Note that for participants with only four interactions, the results for Round 5 are identical to those of Round 4.}
\label{table:overall_model_comparison}

\end{table}

\begin{table}[H]
\centering
\begin{subtable}[t]{\columnwidth}
\centering
\resizebox{1.\columnwidth}{!}{ 
\begin{tabular}{cccccc}
\toprule[2pt]
\diagbox{$X$}{Round 2}{$Y$}& \textit{Persona} & \textit{Generic} & \textit{SBU} & \textit{$H\!M_1$} & \textit{$H\!M_2$} \\ \midrule[2pt]
\textit{Persona} & -- & $0.044$ & $0.047$ & $2.408 \times 10^{-6}$ & $5.730 \times 10^{-5}$ \\ [1ex]
\textit{Generic}   & $0.956$    & --  & $0.187$ &  $8.210 \times 10^{-6}$ & $1.907 \times 10^{-4}$\\ [1ex]
\textit{SBU}    &  $0.953$  & $0.813$  & -- & $4.685 \times 10^{-5}$  & $5.659 \times 10^{-4}$ \\ [1ex]
\textit{$H\!M_1$}     & $1$   & $1$  & $1$  & -- & $0.908$ \\ [1ex]
\textit{$H\!M_2$}   &  $1$   & $1$  & $0.999$ & $0.092$ & --  \\ \bottomrule[2pt]
\end{tabular}
}
\caption{The $p$-values that $X$ outperforms $Y$ in Round 2.}
\label{table:round2_p_value}
\end{subtable}

\vspace{0.2em}

\begin{subtable}[t]{\columnwidth}
\centering
\resizebox{1.\columnwidth}{!}{ 
\begin{tabular}{cccccc}
\toprule[2pt]
\diagbox{$X$}{Round 3}{$Y$}& \textit{Persona} & \textit{Generic} & \textit{SBU} & \textit{$H\!M_1$} & \textit{$H\!M_2$} \\ \midrule[2pt]
\textit{Persona} & -- & $0.004$ & $0.004$ & $3.254 \times 10^{-4}$ & $0.002$ \\ [1ex]
\textit{Generic}   & $0.996$    & --  & $0.231$ &  $0.003$ & $0.021$\\ [1ex]
\textit{SBU}    &  $0.997$  & $0.769$  & -- & $0.005$  & $0.026$ \\ [1ex]
\textit{$H\!M_1$}     & $1$   & $0.997$  & $0.995$  & -- & $0.776$ \\ [1ex]
\textit{$H\!M_2$}   &  $0.998$   & $0.979$  & $0.974$ & $0.224$ & --  \\ \bottomrule[2pt]
\end{tabular}
}
\caption{The $p$-values that $X$ outperforms $Y$ in Round 3.}
\label{table:round3_p_value}
\end{subtable}

\vspace{0.2em}

\begin{subtable}[t]{\columnwidth}
\centering
\resizebox{1.\columnwidth}{!}{ 
\begin{tabular}{cccccc}
\toprule[2pt]
\diagbox{$X$}{Round 4}{$Y$}& \textit{Persona} & \textit{Generic} & \textit{SBU} & \textit{$H\!M_1$} & \textit{$H\!M_2$} \\ \midrule[2pt]
\textit{Persona} & -- & $0.010$ & $0.047$ & $3.760 \times 10^{-5}$ & $0.006$ \\ [1ex]
\textit{Generic}   & $0.990$    & --  & $0.748$ &  $6.182 \times 10^{-4}$ & $0.030$\\ [1ex]
\textit{SBU}    &  $0.953$  & $0.252$  & -- & $5.905 \times 10^{-4}$  & $0.022$ \\ [1ex]
\textit{$H\!M_1$}     & $1$   & $1$  & $1$  & -- & $0.916$ \\ [1ex]
\textit{$H\!M_2$}   &  $0.994$   & $0.969$  & $0.978$ & $0.084$ & --  \\ \bottomrule[2pt]
\end{tabular}
}
\caption{The $p$-values that $X$ outperforms $Y$ in Round 4.}
\label{table:round4_p_value}
\end{subtable}

\vspace{0.2em}
\begin{subtable}[t]{\columnwidth}
\centering
\resizebox{1.\columnwidth}{!}{ 
\begin{tabular}{cccccc}
\toprule[2pt]
\diagbox{$X$}{Round 5}{$Y$}& \textit{Persona} & \textit{Generic} & \textit{SBU} & \textit{$H\!M_1$} & \textit{$H\!M_2$} \\ \midrule[2pt]
\textit{Persona} & -- & $0.640$ & $0.426$ & $0.006$ & $0.001$ \\ [1ex]
\textit{Generic}   & $0.360$    & --  & $0.153$ &  $0.003$ & $5.790 \times 10^{-4}$\\ [1ex]
\textit{SBU}    &  $0.574$  & $0.867$  & -- & $0.007$  & $0.001$ \\ [1ex]
\textit{$H\!M_1$}     & $0.994$   & $0.997$  & $0.993$  & -- & $0.225$ \\ [1ex]
\textit{$H\!M_2$}   &  $0.999$   & $0.999$  & $0.999$ & $0.775$ & --  \\ \bottomrule[2pt]
\end{tabular}
}
\caption{The $p$-values that $X$ outperforms $Y$ in Round 5.}
\label{table:round5_p_value}
\end{subtable}

\caption{The $p$-values from paired Student’s t-tests assessing the
hypothesis that $X$ outperforms $Y$ in round $k$ ($k = 2,3,4,5$) in Experiment 2.1}
\label{table:all_rounds_p_value}
\end{table}

\begin{table}[H]
\centering
\resizebox{1.\columnwidth}{!}{ 
\begin{tabular}{lcc}
\toprule[2pt]
& Group A (Four rounds) & Group B (Five rounds) \\
\midrule[2pt]
$p_{1,2}$ & $4.199 \times 10^{-15}$ & $1.774 \times 10^{-20}$ \\
$p_{2,3}$ & $3.412 \times 10^{-5}$ & $0.005$ \\
$p_{3,4}$ & $0.02$ & $0.016$ \\
$p_{4,5}$ & -- & $0.043$ \\
\bottomrule[2pt]
\end{tabular}
}
\caption{The $p$-values of comparing confidence values between interaction rounds. Specifically, $p_{i,j}$ indicates the $p$-value for the hypothesis that the confidence increases from round $i$ to round $j$. }
\label{table:t-test_confidence}
\end{table}

\begin{table}[H]
\centering \small
\begin{tabular}{cc}
\toprule[2pt]
                               & All Participants \\ \midrule[2pt]
Comprehension Score (out of 5) & 3.32             \\ [1ex]
Satisfaction Score (out of 5)  & 3.12             \\ \bottomrule[2pt]
\end{tabular}
\caption{Comprehension score and satisfaction score.}
\label{table:comprehension_and_satisfaction}
\end{table}

\end{document}